\documentclass[conference]{IEEEtran}
\IEEEoverridecommandlockouts
\usepackage{cite}
\usepackage{amsmath,amssymb,amsfonts}
\usepackage{algorithmic}
\usepackage{graphicx}
\usepackage{textcomp}
\usepackage{xcolor}
\def\BibTeX{{\rm B\kern-.05em{\sc i\kern-.025em b}\kern-.08em
    T\kern-.1667em\lower.7ex\hbox{E}\kern-.125emX}}

\begin{document}

\title{A Real-Time DETR Approach to Bangladesh Road Object Detection for Autonomous Vehicles
}

\author{
\IEEEauthorblockN{Irfan Nafiz Shahan}
\IEEEauthorblockA{\textit{Department of Electrical and Electronic Engineering} \\
\textit{Shahjalal University of Science and Technology}\\
Sylhet, Bangladesh \\
irfannafizislive@gmail.com}
\and
\IEEEauthorblockN{Arban Hossain}
\IEEEauthorblockA{\textit{Department of Robotics and Mechatronics Engineering} \\
\textit{Dhaka University}\\
Dhaka, Bangladesh \\
arbanhossain@gmail.com}
\and
\IEEEauthorblockN{Saadman Sakib}
\IEEEauthorblockA{\textit{Department of Robotics and Mechatronics Engineering} \\
\textit{Dhaka University}\\
Dhaka, Bangladesh \\
saadman.sakib2020@gmail.com}
\and
\IEEEauthorblockN{Al-Mubin Nabil}
\IEEEauthorblockA{\textit{Department of Computer Science and Engineering} \\
\textit{Shahjalal University of Science and Technology}\\
Sylhet, Bangladesh \\
almubinnabil@gmail.com}
}

\maketitle

\begin{abstract}
In the recent years, we have witnessed a paradigm shift in the field of Computer Vision, with the forthcoming of the transformer architecture. Detection Transformers has become a state of the art solution to object detection and is a potential candidate for Road Object Detection in Autonomous Vehicles. Despite the abundance of object detection schemes, real-time DETR models are shown to perform significantly better on inference times, with minimal loss of accuracy and performance. In our work, we used Real-Time DETR (RTDETR) object detection on the BadODD Road Object Detection dataset based in Bangladesh, and performed necessary experimentation and testing. Our results gave a mAP50 score of 0.41518 in the public 60\% test set, and 0.28194 in the private 40\% test set. 

\end{abstract}


\section{Introduction}
Computer vision (CV) plays a critical role in the modern day, and used commonly in some security cameras. Beyond security, the scope of computer vision extends into the realm of autonomous vehicles, where it plays a pivotal role in vehicle detection and tracking within the sensor fusion framework.

For autonomous vehicles, accurate and swift object detection can be a matter of life and death - for proper decision-making depends solely upon the inference of the machine learning model involved that analyses data to provide powerful conclusions to a complex road scenario. Therefore, advances in road vehicle detection is paramount for the future safety and capabilities of autonomous vehicles. 

The forthcoming of major developments in machine learning, such as transformer models,\cite{vaswani2023attention} paved the way towards improved object detection models, namely Detection Transformers (DETRs) \cite{carion2020endtoend}. DETRs took the spotlite in 2020, becoming a cornerstone to faster, lower memory and lower power methods of object detection. Traditionally before the emergence of DETR, Region-based Convolutional Neural Networks (R-CNN) were - and still are - actively being used for making complex computation models. A very popular such architecture is You-Only-Look-Once (YOLO)\cite{Jocher_Ultralytics_YOLO_2023}that utilizes Fast R-CNNs to enable incredible inference times, with it's pretrained model capable of being fine-tuned to significantly commendable accuracy in most scenarios. 

In a  paper by Lv et al\cite{lv2023detrs}, it is shown that Real-Time DETR (RTDETR), outperforms classical YOLO\cite{Jocher_Ultralytics_YOLO_2023} variants and Fast R-CNN models significantly due to it's lower end-to-end latency and customizability. Furthermore does not suffer from any major reduction in accuracy and performances.  RTDETR (Real-Time DETR) represents a fusion of the speed and efficiency of YOLO-style architectures with the expressive power of transformers. Unlike traditional convolutional neural networks (CNNs) used in YOLO\cite{Jocher_Ultralytics_YOLO_2023} and R-CNN variants, transformers rely on self-attention mechanisms to capture global dependencies in the input data, enabling more effective object detection. 

Hence, beyond using conventional YOLOv8\cite{Jocher_Ultralytics_YOLO_2023} based fine-tuned models, we aimed at using RTDETR on the BadODD - Bangladeshi Autonomous Driving Object Detection Dataset \cite{baig2024badodd}. 

\section{Methodology}
\subsection{Understanding BadODD}
\label{understandingbadodd}

BadODD dataset consists of a total of 13 classes, including auto-rickshaw, bicycle, bus, car, cart-vehicle, construction-vehicle, motorbike, person, priority-vehicle, three-wheeler, train, truck, and wheelchair.

\begin{figure}
    \centering
    \includegraphics[width=0.9\linewidth]{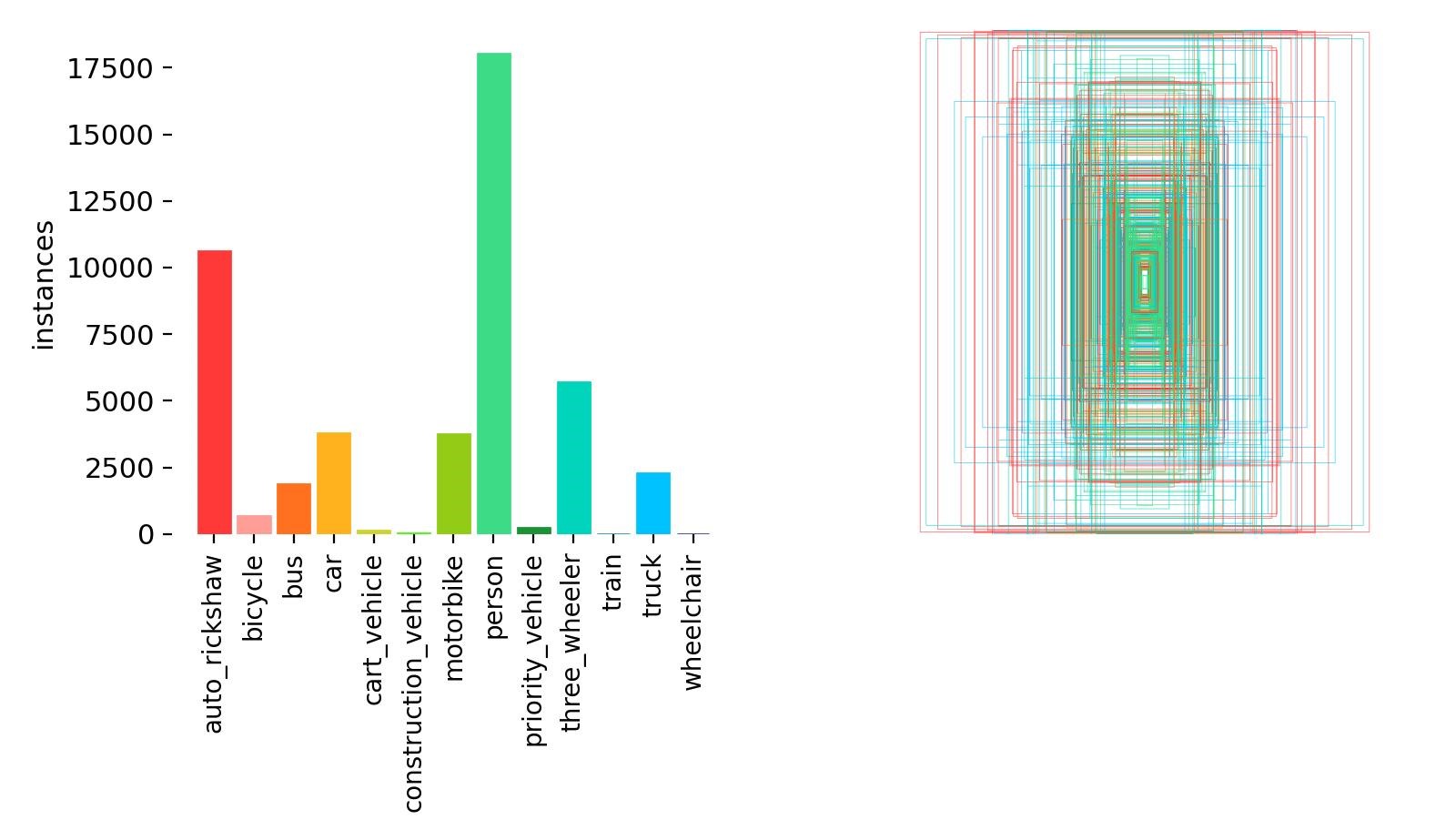}
    \caption{Dataset Analysis of BadODD, showcasing the disparity in model class occurrences and the uneven distribution of bounding box sizes,}
    \label{fig:label_cropped}
\end{figure}

The distribution of classes within the dataset is disproportionate,\ref{fig:label_cropped} with some classes having significantly more instances than others. This class imbalance poses a challenge for training and evaluating object detection models, as it can affect the model's ability to accurately detect and classify objects across all classes.

In the dataset, there are a total of 5896 images available for training and 1964 images for testing. The disparity in class frequencies within these images further exacerbates the challenge of developing a robust object detection model that performs well across all classes.

Beyond these issues, there are some potential concerns in the dataset that pose challenges in training the model.

\begin{enumerate}
    \item \textbf{Flares:} Flares pose a challenge as they may obscure objects in images, potentially reducing detection accuracy. While their removal could improve accuracy, it requires additional resources. \ref{fig:flare_final} We used the Flare7k model to conduct tests. \cite{dai2022flare7k}\cite{dai2023flare7kpp}

    \begin{figure}
        \centering
        \includegraphics[width=0.8\linewidth]{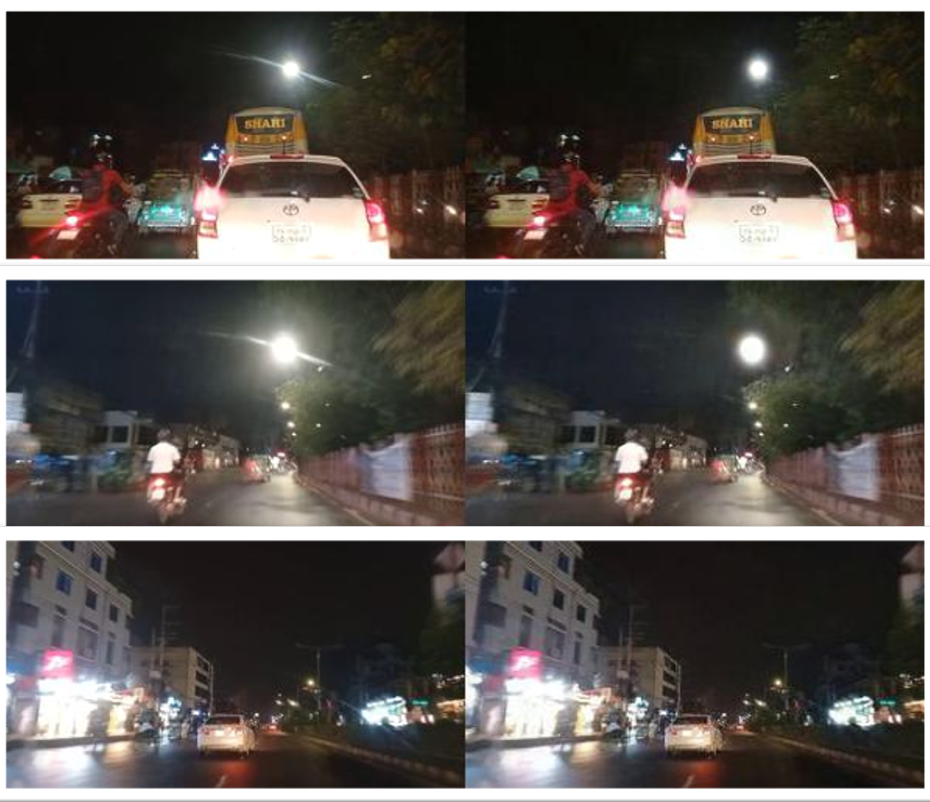}
        \caption{Original picture with flare (left), flare-reduced picture (right)}
        \label{fig:flare_final}
    \end{figure}
    
    \item \textbf{Night Images:} Night images are challenging due to low visibility and sensor artifacts. They often lack clarity, making feature detection difficult. \ref{fig:night_final}. Tested using Dai's Night to Day nighttime model. \cite{dai2023nighttime}

     \begin{figure}
        \centering
        \includegraphics[width=0.9\linewidth]{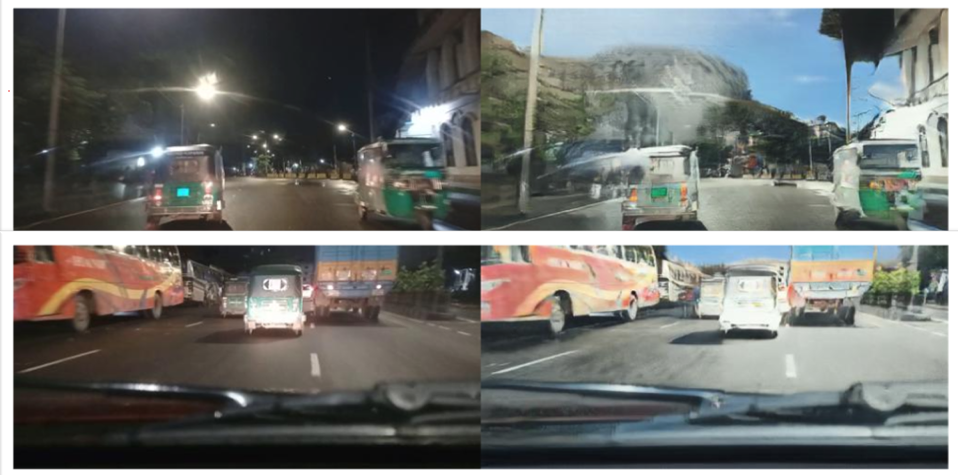}
        \caption{Original night picture (left), night to day converted picture (right)}
        \label{fig:night_final}
    \end{figure}
    
    \item \textbf{Windshield Stains:} Stains on windshields can hinder accurate labeling and bounding box detection, affecting model performance. \ref{fig:stain_final} Blur removal conducted using Wei's reflection model. \cite{wei2019single}

    \begin{figure}
        \centering
        \includegraphics[width=0.9\linewidth]{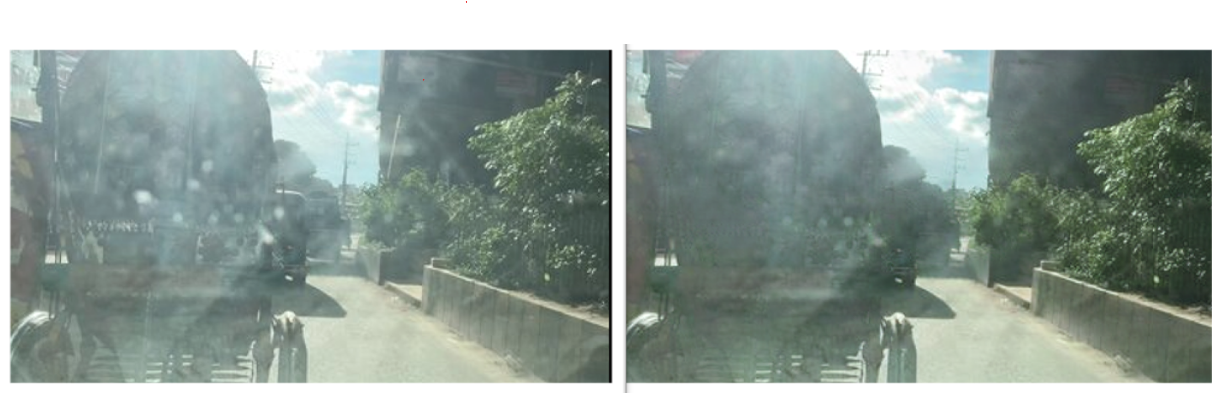}
        \caption{Original picture with windshield stain (left), corresponding stain-reduced image (right)}
        \label{fig:stain_final}
    \end{figure}
    
    \item \textbf{Motion Blur:} Motion blur distorts object features, complicating detection. While deblurring techniques exist, they're computationally intensive.
    
    \item \textbf{Double Inference Pass:} Running inference twice on an image can refine detections but increases computational cost. Optimizing this process is crucial for real-time applications.
\end{enumerate}

We conducted experiments to address these challenges, however, due to constraints in resources and time, these preprocessing methods were not ultimately integrated into our RTDETR model. Nevertheless, our experimentation highlights their relevance and potential for future work. In subsequent iterations, addressing these challenges could enhance the robustness and accuracy of object detection systems, warranting further investigation.

\subsection{Model Selection}
We opted for a Real-Time Detection Transformer (RT-DETR) instead of the regular YOLOv8 pre-trained model. YOLO\cite{Jocher_Ultralytics_YOLO_2023} is reputed for its fast inference time, which often comes with the cost of reduced accuracy. DEtection TRansformers (DETR), first introduced in 2020, leverages vision transformers and their encoder-decoder architecture to predict all objects at once. This approach is simpler, more efficient than regular object detectors, and performs better on state-of-the-art baseline datasets like COCO. But it is also very slow, making it unusable for real-time inference.

RT-DETR mitigates this challenge by adapting and supporting flexible adjustment of inference speed using different decoder layers, removing the need for retraining. This allows us to take advantage of the strength of Vision Transformers while also retaining a very high inference speed. Transformers also enable parallel processing and capture long-range dependencies better.

\subsection{Data Preprocessing} \label{preprocess}
Our workflow includes several layers of significant preprocessing to ensure valid and consistent input to our model. The RT-DETR architecture expects $416 \times 416$ pixel images. We resized the images and adjusted the labels accordingly. We applied augmentation such as blur, median blur, and grayscale conversion. These allow the model to generalize better over different types of inputs. We also tested other augmentations such as perspective transform, random scaling, random translation, and mosaic augmentation.

\subsection{Model Configuration and Hyperparameter Tuning}
We considered multiple configurations for our RT-DETR model. These configurations are listed in Table. For optimal tuning, we considered whether the model was pre-trained or not, the learning rate, warmup iterations, weight decay, momentum, and image batch size.


\begin{table}[h!]
    \centering
    \begin{tabular}{|c|c|}
         \hline
         learning rate & 0.001 \\
         \hline
         warmup iterations & 3 \\
         \hline
         momentum & 0.937 \\
         \hline
         epoch & 50 \\
         \hline
         image batch size & 16 \\
         \hline
    \end{tabular}
    \vspace{1em}
    \caption{Tuned Hyperparameters}
    \label{tab:my_label}
\end{table}

\subsection{Training}
After selecting the model and performing preprocessing, we proceeded with the training process to develop the deep learning model for object detection. Optimization was achieved using the AdamW optimizer, and the hyperparameters listed in table \ref{tab:my_label}. We used a slow learning rate and a moderate number of epoch to reach a appreciable mAP score.

\section{Results and Discussion}
\begin{figure*}[h!]
    \centering
    \includegraphics[width=.75\textwidth]{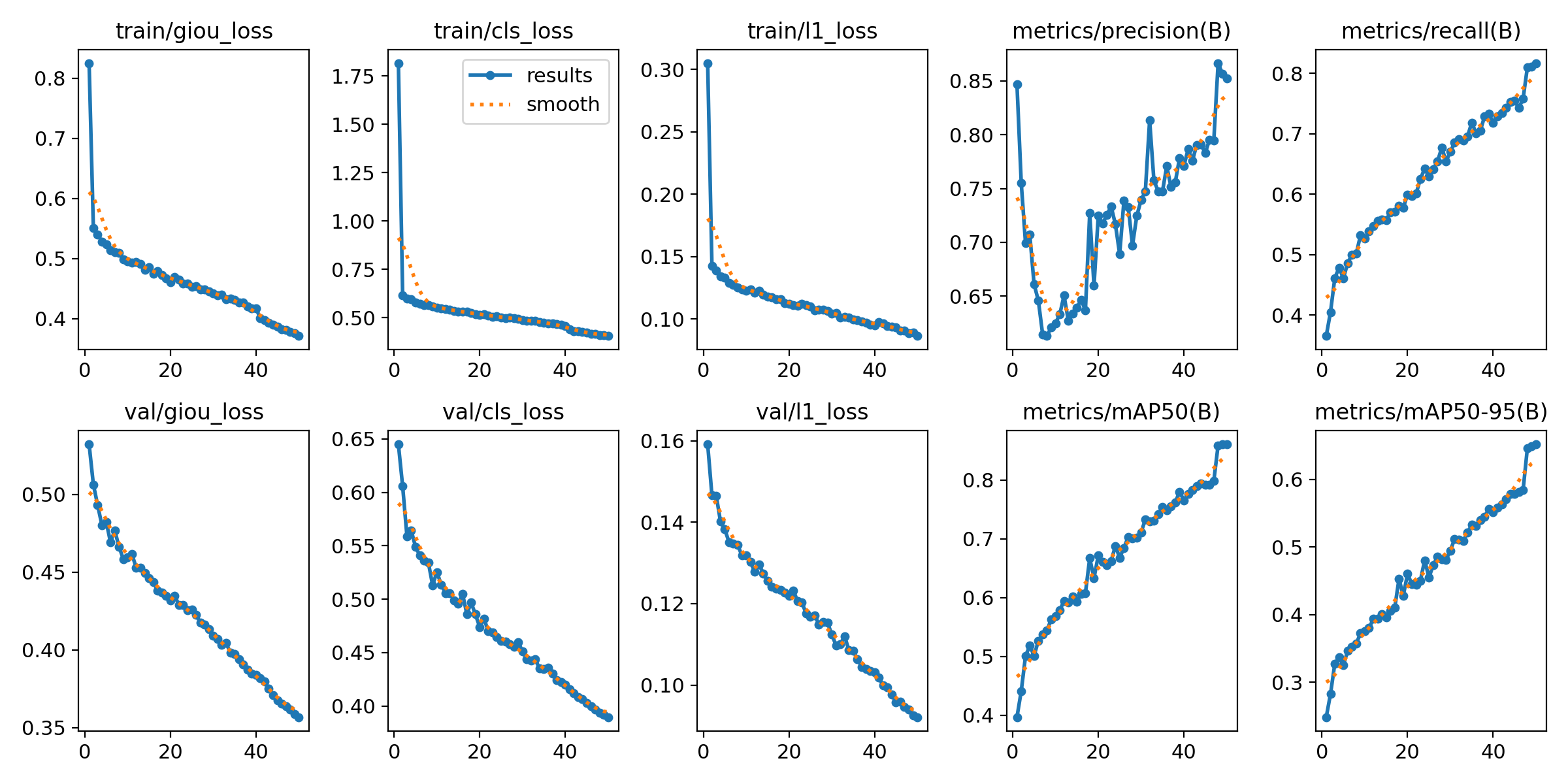}
    \caption{Performance Metrics for the RTDETR-X model}
    \label{fig:loss_curves}
\end{figure*}

The performance metrics for the model are illustrated in table \ref{tab:results}. The model achieves a mean Average Precision (mAP) score of 0.4151 on the public 60\% of the dataset. Accordingly, it scores 0.2891 on the remaining 40\%. Table \ref{tab:results} shows that the model achieves high precision and recall values. The final loss of the model settles at 0.000808. The loss curves are also presented in the figure below (Figure \ref{fig:loss_curves}). The model also achieves an average inference time of 22.44ms (Table \ref{tab:speed}). These results underscore the model's ability to generalize well across diverse data distributions.

Table \ref{tab:results} further highlights the model's impressive precision and recall values, indicating its capacity to achieve high accuracy while minimizing false positives and negatives. This balanced performance is crucial for real-world applications where reliable object detection is paramount.

\begin{table}[h!]
    \centering
    \begin{tabular}{|c|c|}
         \hline
         preprocess & 0.2 ms\\
         \hline
         inference & 22.4ms \\
         \hline
         loss & 0.0ms \\
         \hline
         postprocess & 0.7ms\\
         \hline
    \end{tabular}
    \vspace{1em}
    \caption{Time Taken Per Image}
    \label{tab:speed}
\end{table}

The model's average inference time of 22.44 milliseconds, as depicted in Table \ref{tab:speed}, is noteworthy. Low inference times are critical for real-time applications such as autonomous driving and surveillance systems, where rapid decision-making is essential for ensuring safety and efficiency. The model's efficient inference time enables timely processing of input data, facilitating swift responses to dynamic environmental changes.

Despite its notable achievements, the model may encounter certain challenges and limitations. For instance, while it excels in detecting objects of various sizes, it may struggle with accurately identifying extremely small or occluded objects.

\begin{figure}[h!]
\centering
\begin{minipage}[b]{0.21\textwidth}
\centering
\includegraphics[width=\linewidth]{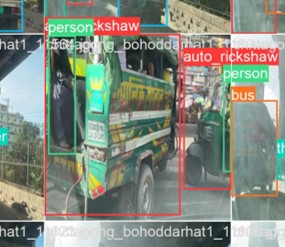}
\caption{Ground Truth Labels showing congested persons}
\label{fig:conj_label}
\end{minipage}
\hfill
\begin{minipage}[b]{0.21\textwidth}
\centering
\includegraphics[width=\linewidth]{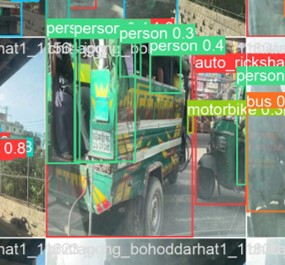}
\caption{Predicted Labels Trying to Identify the congested labels}
\label{fig:conj_pred}
\end{minipage}
\end{figure}

Addressing such challenges requires ongoing research and development efforts aimed at enhancing the model's robustness and adaptability across diverse scenarios.

\begin{table*}[h!]
    \centering
    \begin{tabular}{|r|r|r|r|r|r|r|}
\hline
        Class  &  Images  & Instances  &    Box(P) &          Box(R) &      Box(mAP50) & Box(mAP50-95)\\
\hline
 { all } & 5896 & 47118 & 0.852 & 0.815 & 0.861 & 0.652 \\
 { auto rickshaw } & 5896 & 10614 & 0.931 & 0.939 & 0.968 & 0.759 \\
 { bicycle } & 5896 & 673 & 0.9 & 0.892 & 0.933 & 0.592 \\
 { bus } & 5896 & 1885 & 0.954 & 0.936 & 0.966 & 0.74 \\
 { car } & 5896 & 3785 & 0.95 & 0.914 & 0.953 & 0.724 \\
 { construction vehicle } & 5896 & 141 & 0.885 & 0.879 & 0.916 & 0.685 \\
 { motorbike } & 5896 & 3749 & 0.916 & 0.92 & 0.944 & 0.613 \\
 { person } & 5896 & 18010 & 0.9 & 0.858 & 0.915 & 0.599 \\
 { priority vehicle } & 5896 & 229 & 0.936 & 0.953 & 0.973 & 0.792 \\
 { three wheeler } & 5896 & 5710 & 0.927 & 0.946 & 0.971 & 0.762 \\
 { train } & 5896 & 1 & 0 & 0.916 & 0.923 & 0.705 \\
 { truck } & 5896 & 2296 & 0.971 & 0.948 & 0.98 & 0.752 \\
 { wheelchair } & 5896 & 2 & 1 & 0.5 & 0.75 & 0.75\\
    \hline
    \end{tabular}
    \vspace{1em}
    \caption{Results tabulated for each class.}
    \label{tab:results}
\end{table*}

\begin{figure}[h!]
    \centering
    \includegraphics[width=.5\textwidth]{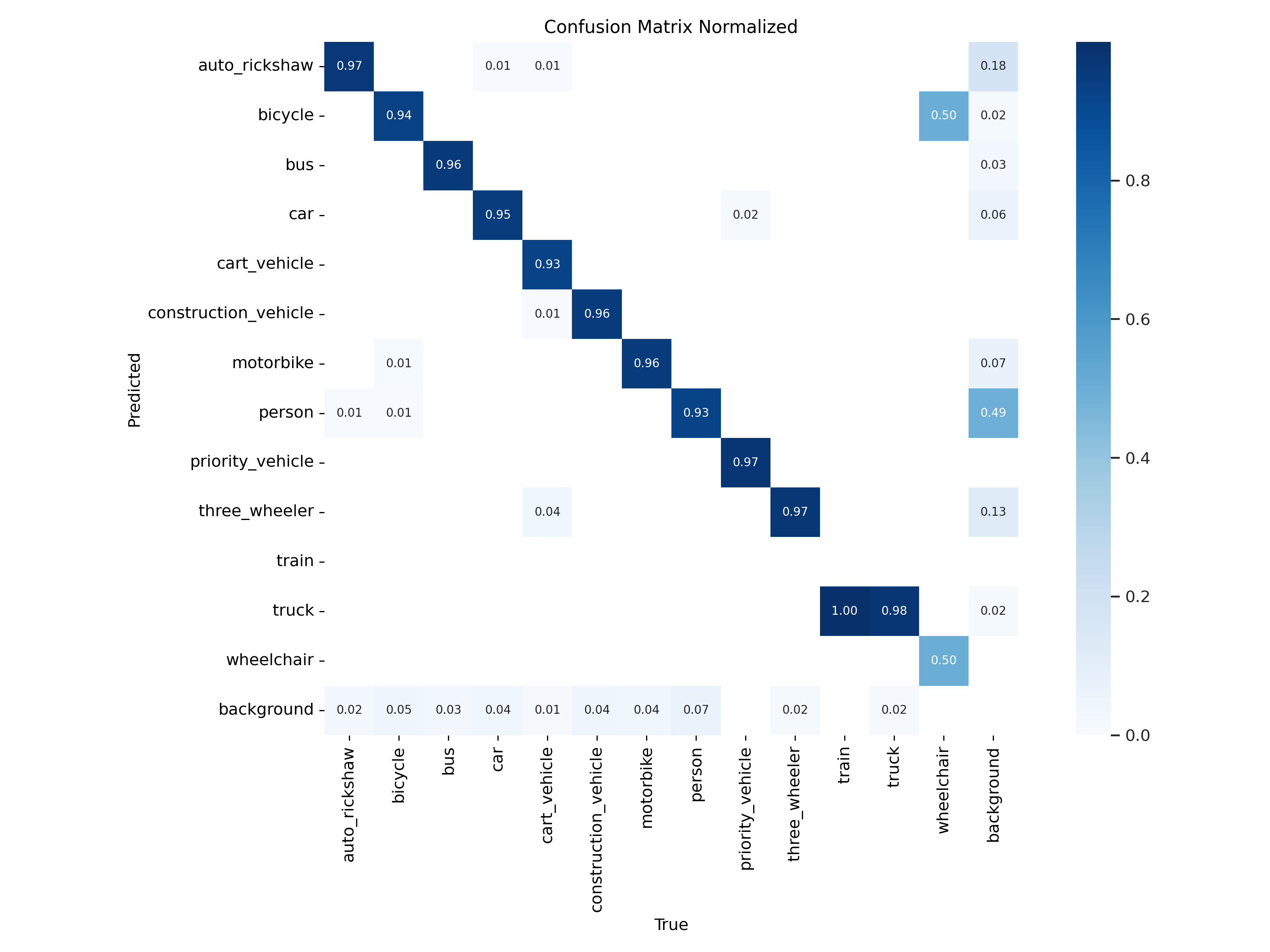}
    \vspace{1em}
    \caption{Confusion Matrix for each of the categories}
    \label{fig:enter-label}
\end{figure}





\section{Conclusion and Future Work}
In this work, we propose a real-time object detection approach using a fine-tuned RT-DETR architecture to detect road objects in Bangladesh. While our method addresses key challenges in the domain, further improvements are needed. We explored several augmentation techniques, which show promise, and suggest that enhancing preprocessing to handle blurs, flares, glass stains, and night images could lead to significant improvements, along with new challenges. Additionally, the dataset is imbalanced, with only a few images of classes like \textit{wheelchair} and \textit{train}, which affects model performance. Lastly, the model struggles with congested areas, where it erroneously predicts multiple objects in a single space, an issue worth further investigation.

\vspace{12pt}


\begin{thebibliography}{00}
\bibitem{vaswani2023attention}
Ashish Vaswani, Noam Shazeer, Niki Parmar, Jakob Uszkoreit, Llion Jones, Aidan N. Gomez, Lukasz Kaiser, Illia Polosukhin. 
"Attention Is All You Need". arXiv preprint arXiv:1706.03762, 2023.

\bibitem{carion2020endtoend}
Nicolas Carion, Francisco Massa, Gabriel Synnaeve, Nicolas Usunier, Alexander Kirillov, Sergey Zagoruyko. 
"End-to-End Object Detection with Transformers". arXiv preprint arXiv:2005.12872, 2020.

\bibitem{lv2023detrs}
Wenyu Lv, Yian Zhao, Shangliang Xu, Jinman Wei, Guanzhong Wang, Cheng Cui, Yuning Du, Qingqing Dang, Yi Liu. 
"DETRs Beat YOLOs on Real-time Object Detection". arXiv preprint arXiv:2304.08069, 2023.

\bibitem{Jocher_Ultralytics_YOLO_2023}
Glenn Jocher, Ayush Chaurasia, Jing Qiu. 
"Ultralytics YOLO". 
GitHub repository, version 8.0.0, 2023. 

\bibitem{baig2024badodd} 
Mirza Nihal Baig, Rony Hajong, Mahdi Murshed Patwary, Mohammad Shahidur Rahman, Husne Ara Chowdhury. 
"BadODD: Bangladeshi Autonomous Driving Object Detection Dataset". arXiv preprint arXiv:2401.10659, 2024.





\bibitem{wei2019single}
Kaixuan Wei, Jiaolong Yang, Ying Fu, David Wipf, Hua Huang. 
"Single Image Reflection Removal Exploiting Misaligned Training Data and Network Enhancements". 
IEEE Conference on Computer Vision and Pattern Recognition, 2019.

\bibitem{dai2022flare7k}
Yuekun Dai, Chongyi Li, Shangchen Zhou, Ruicheng Feng, Chen Change Loy. 
"Flare7K: A Phenomenological Nighttime Flare Removal Dataset". 
Thirty-sixth Conference on Neural Information Processing Systems Datasets and Benchmarks Track, 2022.

\bibitem{dai2023flare7kpp}
Yuekun Dai, Chongyi Li, Shangchen Zhou, Ruicheng Feng, Yihang Luo, Chen Change Loy. 
"Flare7K++: Mixing Synthetic and Real Datasets for Nighttime Flare Removal and Beyond". 2023.

\bibitem{dai2023nighttime}
Yuekun Dai, Yihang Luo, Shangchen Zhou, Chongyi Li, Chen Change Loy. 
"Nighttime Smartphone Reflective Flare Removal using Optical Center Symmetry Prior". 
Proceedings of the IEEE Conference on Computer Vision and Pattern Recognition (CVPR), 2023.


\bibitem{kwak2021adverse}
Jeong-gi Kwak, Youngsaeng Jin, Yuanming Li, Dongsik Yoon, Donghyeon Kim, Hanseok Ko. 
"Adverse weather image translation with asymmetric and uncertainty-aware GAN". 
arXiv preprint arXiv:2112.04283, 2021.



\end{thebibliography}
\end{document}